\documentclass[journal]{IEEEtran}

\usepackage{graphicx}
\usepackage{multirow}
\usepackage{amssymb}
\usepackage{booktabs}

\usepackage[
pagebackref=true,
breaklinks=true,
letterpaper=true,
colorlinks,
bookmarks=false
]{hyperref}

\usepackage{cite}
\usepackage{amsmath}

\begin{document}

\title{Learning Category- and Instance-Aware Pixel Embedding for Fast Panoptic Segmentation}

\author{
	Naiyu~Gao,~\IEEEmembership{Student Member,~IEEE,}
  Yanhu~Shan,~\IEEEmembership{Member,~IEEE,}
  Xin~Zhao,~\IEEEmembership{Member,~IEEE,}
  and~Kaiqi~Huang,~\IEEEmembership{Senior~Member,~IEEE}
 \thanks{This work is supported in part by the National Natural Science Foundation of China (Grant No. 61721004 and No.61876181), the Projects of Chinese Academy of Science (Grant No. QYZDB-SSW-JSC006), and the Youth Innovation Promotion Association CAS. (\emph{Corresponding author:} Xin Zhao.)}
 \thanks{Naiyu Gao and Xin Zhao are with the Center for Research on Intelligent System and Engineering (CRISE), Institute of Automation, Chinese Academy of Sciences (CASIA), Beijing 100190, China, and also with the School of Artificial Intelligence, University of Chinese Academy of Sciences (UCAS), Beijing 100049, China (e-mail: gaonaiyu2017@ia.ac.cn; xzhao@nlpr.ia.ac.cn).}
 \thanks{Yanhu Shan is with Horizon Robotics, Beijing 100000, China (e-mail: yanhu.shan@horizon.ai).}
 \thanks{Kaiqi Huang is with the Center for Research on Intelligent System and Engineering (CRISE), Institute of Automation, Chinese Academy of Sciences (CASIA), Beijing 100190, China, the School of Artificial Intelligence, University of Chinese Academy of Sciences, Beijing 100049, P.R.China, and also with the CAS Center for Excellence in Brain Science and Intelligence Technology, Shanghai 200031, P.R.China. E-mail: kqhuang@nlpr.ia.ac.cn. }
}

\markboth{IEEE TRANSACTIONS ON IMAGE PROCESSING}
{Shell \MakeLowercase{\textit{et al.}}: Bare Demo of IEEEtran.cls for IEEE Journals}
\maketitle

\begin{abstract}
Panoptic segmentation (PS) is a complex scene understanding task that requires providing high-quality segmentation for both \emph{thing} objects and \emph{stuff} regions. Previous methods handle these two classes with semantic and instance segmentation modules separately, following with heuristic fusion or additional modules to resolve the conflicts between the two outputs. This work simplifies this pipeline of PS by consistently modeling the two classes with a novel PS framework, which extends a detection model with an extra module to predict category- and instance-aware pixel embedding (CIAE). CIAE is a novel pixel-wise embedding feature that encodes both semantic-classification and instance-distinction information. At the inference process, PS results are simply derived by assigning each pixel to a detected instance or a \emph{stuff} class according to the learned embedding. Our method not only demonstrates fast inference speed but also the first one-stage method to achieve comparable performance to two-stage methods on the challenging COCO benchmark.
\end{abstract}


\section{Introduction}
\IEEEPARstart{S}{cene} parsing is an important vision task and serves as the basis of many practical applications, including autonomous driving, robotics, and image editing. The recently proposed panoptic segmentation (PS) task~\cite{panoptic} requires determining the semantic category of each pixel in an input image while identifying and segmenting each object instance.
Moreover, PS segments both amorphous uncountable regions (\emph{stuff}) and countable objects (\emph{thing}), which provides more complete scene information when comparing with semantic segmentation or instance segmentation.

A straight forward solution for panoptic segmentation is to deal \emph{stuff} and \emph{thing} classes with semantic segmentation and instance segmentation branches, respectively, as in the existing methods~\cite{panopticFPN,Liu_2019_CVPR,Li_2019_CVPR,xiong19upsnet}. 
The semantic segmentation branch outputs pixel-level semantic classification, while the instance segmentation branch detects instances and produces pixel-level instance-distinction. After that, the outputs from two branches are fused to get the final PS results, where heuristic rules or additional modules are developed to handle the conflicts between the two outputs. However, this complex pipeline brings heavy computation complexity and memory footprint, especially for the two-stage methods, where time-consuming instance segmentation modules are employed. Recently, one-stage methods~\cite{arxiv_2019_yang_deeperlab,SSAP_Gao_ICCV,SSAP_Gao_CSVT,PCV_Wang_2020_CVPR,cheng2020panoptic,real-time-panoptic} are developed, where the instance-distinction information is provided with pixel-level instance-sensitive features. Although having advantages in efficiency, the existing one-stage methods fail to achieve comparable performance to two-stage approaches.

\begin{figure}
  \begin{center}
  \includegraphics[width=1.\linewidth]{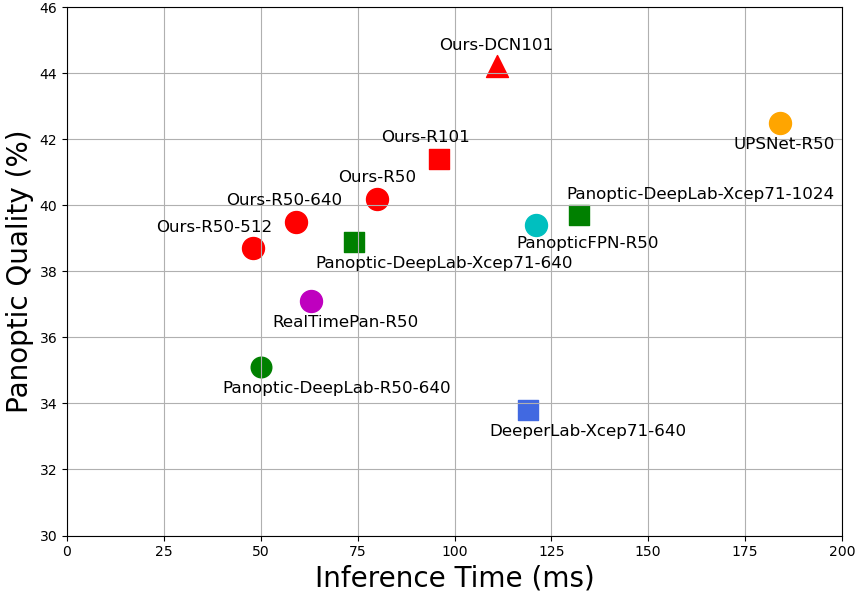}\\
  \end{center}
  \vspace{-.8em}
\caption{{\bf Speed-accuracy trade-off} on the COCO validation set for recent panoptic segmentation methods. Inference times of our method, Panoptic FPN~\cite{panopticFPN}, and UPSNet~\cite{xiong19upsnet} are tested with NVIDIA 2080Ti GPU. Inference times of other methods\cite{arxiv_2019_yang_deeperlab,real-time-panoptic,cheng2020panoptic} are tested with Tesla V100 GPU. Our method shows advantages in both speed and accuracy.}
\label{fig:acc_rt}
\end{figure}

Instead of modeling \emph{thing} and \emph{stuff} classes separately and handling the conflicts at the later fusing process, this work proposes to consistently model the two classes to simplify the whole pipeline. Specifically, we propose a novel PS framework that extends a one-stage detection model with an additional module to predict category- and instance-aware pixel embedding (CIAE), which encodes cues for both pixel-wise semantic-classification and instance-distinction. Different from the original instance-aware pixel embedding (IAE)~\cite{discriminative_loss_2017_cvprw,deep_metric,kong2018recurrent}, CIAE is specifically designed to map each pixel into a feature space where cues about semantic-classifying and instance-differentiating are modeled simultaneously. Concretely, the feature space is divided into category-specific sub-spaces by the learned category-aware query embeddings, and the pixels are mapped not only to distinguish instances but also into corresponding sub-spaces. In this way, the learned pixel embedding captures both category-classifying and instance-distinguishing information.

In the inference procedure, the PS result is simply derived by assigning each pixel to a detected instance or a \emph{stuff} class according to its embedding feature. Specifically, instance proposals and the CIAE embedding map are firstly predicted with the detection and embedding branches, respectively. After that, query embeddings for each instance and \emph{stuff} class are obtained. The former is derived from the predicted embedding map based on where the instance is generated, while the latter is a learned memory embedding. 
Finally, the PS result is obtained by calculating the cosine similarity between pixel embeddings and each query embedding, which is simply implemented as matrix multiplication.

Thanks to the consistency of modeling, our method not only simplifies the PS pipeline but also gains advantages in end-to-end learning and achieves good performance. Experiments on the challenging COCO~\cite{lin2014microsoft} dataset demonstrate the benefits of our method. As shown in Figure~\ref{fig:acc_rt}, our model surpasses all one-stage methods in both accuracy and inference speed. We are also the first one-stage method to achieve comparable performance to two-stage methods on the COCO dataset.

Our contributions are summarized as the following:
\begin{itemize}
\item {This work proposes the concept of category- and instance-aware pixel embedding (CIAE), which captures cues about semantic-classification and instance-distinction simultaneously to consistently model countable \emph{thing} objects and uncountable \emph{stuff} regions. The pipeline of panoptic segmentation is thus much simplified.}
\item{This work demonstrates the feasibility of jointly modeling pixel-level semantic-classification and instance-distinction, instead of employing semantic segmentation and instance segmentation modules separately.}
\item {This work proposes a novel one-stage panoptic segmentation framework by incorporating the proposed CIAE with a one-stage detector. On the COCO benchmark, the proposed model surpasses all one-stage methods in both accuracy and inference speed and is the first one-stage method that achieves comparable performance against time-consuming two-stage methods.}
\end{itemize}

\begin{figure*}
\centering
\includegraphics[width=.9\linewidth]{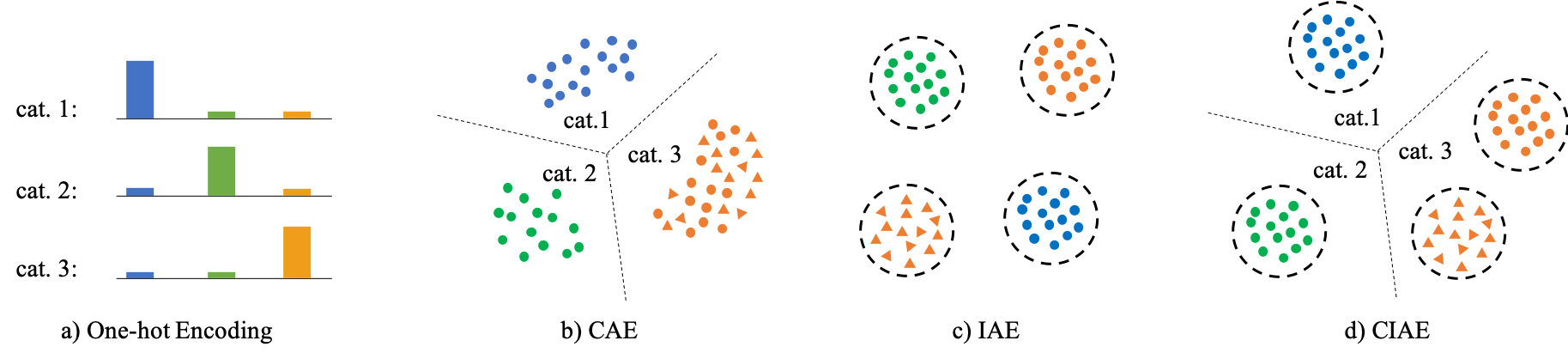}
\vspace{-.5em}
\caption{
{\bf Illustration of different pixel embeddings. }
In b) c) and d), each pixel is mapped into a point in a feature space, and points with the same color are with the same semantic category; points with the same color and shape are from the same instance.
a) One-hot encoding for classification, which can be regarded as a special case of CAE. 
b) CAE (category-aware embedding) maps each pixel into a specific sub-space according to its semantic category. 
c) IAE (instance-aware embedding) maps pixels closer to pixels from the same instance than to pixels from different instances. 
d) CIAE (category- and instance-aware embedding) maps each pixel into the category-specific sub-space while pixels from different instances are distinguishable. 
In practice, the embedding features are distributed on a hypersphere and the cosine similarity is employed.
Best viewed in color and zoom.}
\label{fig:ciae}
\end{figure*}

\section{Related Work}
\subsection{Instance Segmentation}
Instance segmentation requires to detect and segment each object instance in an image. Recent methods can be divided into two categories: two-stage and one-stage methods. The first two-stage instance segmentation method Mask R-CNN~\cite{He_2017_ICCV} extends the two-stage detector Faster R-CNN~\cite{ren2015faster} by adding a mask predicting branch. The following works~\cite{Liu_2018_CVPR,chen2018masklab:,huang2019mask,chen2019hybrid} promote Mask R-CNN with advanced feature aggregation or mask prediction heads. These two-stage methods are showing great performance, but the feature re-pooling and re-sampling operations cause large computational overhead and greatly reduce the efficiency. The one-stage methods are also developed for the instance segmentation task. Some one-stage methods~\cite{xu2019explicit,xie2019polarmask,Bolya_2019_ICCV,Chen_2019_ICCV} follow the detect-then-segment manner and extend one-stage detectors~\cite{redmon2017yolo9000,tian2019fcos:} to predict masks. Other methods design instance-aware features~\cite{dai2016instance-aware,li2017fully,kong2018recurrent,discriminative_loss_2017_cvprw,deep_metric,Liu_2018_ECCV,SSAP_Gao_ICCV,kendall2018multi-task,Neven_2019_CVPR} and obtain instance masks through clustering during post-processing. The instance-aware features can be either explicit instance-aware features~\cite{dai2016instance-aware,li2017fully}, discriminative pixel embeddings~\cite{kong2018recurrent,discriminative_loss_2017_cvprw,deep_metric}, pixel-pair affinities~\cite{Liu_2018_ECCV,SSAP_Gao_ICCV} or instance vectors~\cite{kendall2018multi-task,Neven_2019_CVPR} toward object centroids.

Our method is inspired by the pixel embedding based methods~\cite{kong2018recurrent,discriminative_loss_2017_cvprw,deep_metric} but differs in many aspects. Firstly, the learned pixel embeddings in previous works are only distinguishable to object instances but agnostic to semantic categories. In our method, the learned pixel embedding encodes both instance-distinction and semantic category information. Secondly, this work proposes to learn pixel embeddings with the triplet loss to query embeddings, which is quite different from previous methods~\cite{kong2018recurrent,discriminative_loss_2017_cvprw,deep_metric}. Importantly, time-consuming clustering algorithms like mean-shift~\cite{fukunaga1975the} or K-means~\cite{Arthur2007k} are employed in  previous methods~\cite{kong2018recurrent,discriminative_loss_2017_cvprw,deep_metric} to group pixels. However, in our method, instances are proposed with the detection module and pixels are simply assigned by calculating the similarity between pixel embeddings and each query embedding.

\subsection{Panoptic Segmentation}
Panoptic segmentation requires to assign a semantic label and an instance id for each pixel in an image. Pixels with the same label and id belong to the same object. The instance id is ignored for \emph{stuff} labels. Most state-of-the-art methods employ semantic segmentation and instance segmentation modules to tackle this problem, following with fusion steps to merge the two outputs. OANet~\cite{Liu_2019_CVPR} proposes the Spatial Ranking module to resolve the occlusion between predicted instance masks. Panoptic FPN~\cite{panopticFPN} extends Mask R-CNN~\cite{He_2017_ICCV} with a semantic segmentation branch. AUNet~\cite{Li_2019_CVPR} introduces attention modules to promote the performance of \emph{stuff} classes with the information from the \emph{thing} branch. A parameter-free panoptic head is proposed in UPSNet~\cite{xiong19upsnet} to resolves the conflicts in the fusion process by predicting an extra `unknown' class. Porzi et al.~\cite{porzi2019seamless} propose a segmentation head to integrate the multi-scale features from FPN~\cite{lin2017feature} with contextual information provided by a light-weight, DeepLab-inspired module~\cite{chen2018deeplabv2:}. AdaptIS~\cite{Sofiiuk_2019_ICCV} generates instance masks from point proposals. Recently, OCFusion~\cite{Lazarow_2020_CVPR}, BGRNet~\cite{Wu_2020_CVPR} and BANet~\cite{Chen_2020_CVPR} propose to improve results with instance occlusion estimation and bidirectional feature aggregation. Li et al.~\cite{Li_2020_CVPR} exploit a pairwise instance affinity operation for end-to-end learning. In these methods, the pixel-level semantic classification is provided with semantic segmentation branch as pixel-wise classification scores, and the pixel-level instance-distinction is provided with instance segmentation branch as foreground masks. The instance segmentation branches are mostly two-stage models like Mask R-CNN~\cite{He_2017_ICCV}, which shows good performance but is inefficient and memory-consuming. Heuristic rules or additional models are also required to tackle the conflicts in the fusion process, which limits the accuracy and efficiency of the whole system.

Instead of employing two-stage instance segmentation modules like Mask R-CNN~\cite{He_2017_ICCV}, some other works attempt to solve panoptic segmentation in a one-stage manner, where the instance-distinction information is provided with pixel-level instance-sensitive features. DeeperLab~\cite{arxiv_2019_yang_deeperlab} proposes to solve panoptic segmentation by predicting key-point and multi-range offset heat-maps, following with a grouping process~\cite{Papandreou_2018_ECCV}. SSAP~\cite{SSAP_Gao_ICCV,SSAP_Gao_CSVT} proposes to distinguish instances with pixel-pair affinity~\cite{Liu_2018_ECCV} pyramid and graph-partition method~\cite{keuper2015efficient}. Panoptic-DeepLab~\cite{cheng2020panoptic} predicts instance centers as well as the offset from each pixel to its corresponding center. PCV~\cite{PCV_Wang_2020_CVPR} emerges instances from pixel consensus on centroid locations and group pixels with back-projection. Hou et al.~\cite{real-time-panoptic} reuse the predictions from dense object detection with a global self-attention mechanism. Although showing advantages in efficiency, the existing one-stage methods can not achieve comparable accuracy to the state-of-the-art methods.

Unlike previous methods model pixel-level semantic-classification and instance-distinction separately, this work proposes category- and instance-aware pixel embedding to encode the two pixel-level features jointly and resolve the conflicts of \emph{thing} and \emph{stuff} classes at the modeling stage.

\section{Category- and Instance-Aware Pixel Embedding}
In this section, we show how to jointly model pixel-level semantic-classification and instance-distinction with category- and instance-aware pixel embedding (CIAE) and introduce the CIAE-loss for learning the pixel embeddings.

\subsection{Problem Formulation}
Drawing inspiration from instance-aware pixel embedding (IAE)~\cite{discriminative_loss_2017_cvprw,deep_metric,kong2018recurrent}, our CIAE is specifically designed. The original IAE is learned to distinguish different instances. Concretely, each pixel is mapped to a feature space. According to the learned pixel embedding, pixels belong to the same instance are close to each other and pixels belong to different instances are separated with a large margin, as shown in Figure~\ref{fig:ciae}(c). However, the learned IAE is category-agnostic. Similarly, we refer to pixel embeddings that can determine semantic categories but are ignorant of instances as category-aware pixel embeddings (CAE). As shown in Figure~\ref{fig:ciae}(b), CAE can be realized by mapping pixels into a category-specific sub-space. One-hot encoding for classification can be regarded as a special case of CAE, where the sub-spaces are pre-defined. CIAE is the combination of CAE and IAE, which maps each pixel into the category-specific sub-space while pixels from different instances are distinguishable, as shown in Figure~\ref{fig:ciae}(d). In this way, both semantic classification and object distinction are achieved with a single embedding map.

Given the embedding map $P$, $p_i$ and $p_j$ represent the embeddings for pixels $i$ and $j$, respectively. In practice, the pixel embeddings are distributed on a hypersphere after L2 normalization:
\begin{equation}
	L2\_norm(p_i)=\frac{p_i}{\|p_i\|_2}.
\end{equation}
The distance between the two embeddings is measured with cosine similarity:
\begin{align}
	D_{cosine}(p_i,p_j)&=1-\frac{p_i\cdot p_j}{\|p_i\|\cdot\|p_j\|}.
\end{align}
Assuming the class numbers for \emph{thing} and \emph{stuff} classes are $C^{th}$ and $C^{st}$, respectively. The number of \emph{thing} instances is $N_{th}$. $c_i \in\{0,1,\dots,C^{st}-1,\dots,C^{st}+C^{th}-1\}$ represents the semantic category of pixel $i$. Meanwhile, $s_i\in\{0,1,\dots,C^{st}-1,\dots,C^{st}+N^{th}-1\}$ represents the segment id of pixel $i$. Pixels belong to the same \emph{stuff} class or \emph{thing} instance are with the same segment id. Note that $s_i$ and $c_i$ are the same for pixels of \emph{stuff} classes.

\subsection{CIAE-Loss}
\label{section:embedding_loss}
In this part, the CIAE-loss for learning CIAE is introduced. As mentioned above, learning CIAE is to map each pixel into the category-specific sub-space while pixels from different instances are distinguishable. For this purpose, the proposed CIAE-loss consists of two sub-losses $L_{CAE}$ and $L_{IAE}$, which respectively improve the discrimination of embeddings between categories and instances. Details are introduced below.

Unlike previous works~\cite{discriminative_loss_2017_cvprw,deep_metric,kong2018recurrent} learning IAE embeddings with exact sampled pixels, this work proposes to learn CIAE by optimizing the distances from each pixel embedding to the mean embeddings of segments. Concretely, the mean embeddings are calculated as the mean directions of pixel embeddings in each segment, following with L2 normalization:
\begin{align}
	\label{eq:mean_emb_sem}
	m_{c}^{k}&=L2\_norm\big(\sum_{c_{i}=k}{p_{i}}\big),\\
	\label{eq:mean_emb_ins}
	m_{s}^{k}&=L2\_norm\big(\sum_{s_{i}=k}{p_{i}}\big).
\end{align}
$M_s=\{m_{s}^{k} | k=0,1,\dots,C^{st}+N^{th}-1\}$ and $M_c=\{m_{c}^{k} | k=0,1,\dots,C^{st}+C^{th}-1\}$ are mean embeddings for instance and category id maps, respectively. In order to suppress noise, the mean embeddings are calculated with the embedding map before L2 normalization. 

In order to encode CIAE, each pixel embedding is supposed to be close to the correct category and instance mean embedding simultaneously. To this end, two triplet losses are developed to learn the pixel embeddings with $M_s$ and $M_c$ as query embeddings, respectively. Given query embeddings, each pixel corresponds to one of them. Using $M_c$ as example, $d^i_{pos}$ and $d^{ik}_{neg}$ are defined as:
\begin{align}
	d^i_{pos}    &= D_{cosine}(p_i,m_{c}^{c_{i}}),\\
	d^{ik}_{neg} &= 
	\left\{ 
	\begin{array}{ll}
		D_{cosine}(p_i,m_c^{k}) &\textrm{ if }k\ne c_{i},\\
		0            &\textrm{ if }k=c_{i}.
	\end{array} 
	\right.
\end{align}

Given query embeddings $M$, the embedding loss $L_{emb}$ is obtained by:
\begin{align}
	l^{'ik}_{emb} &= \max\{d^i_{pos}-d^{ik}_{neg}+margin,\; 0\},\\
	l^{ik}_{emb}  &= -\log{(1-\frac{l^{'ik}_{emb}}{2+margin})},\\
	L_{emb}(P, M) &= \frac{1}{N}\sum_{i}\sum_{k}^{|M|}\mathbb{I}[{l^{ik}_{emb}}>\tau_{K}]\cdot {l^{ik}_{emb}}.
\end{align}
$\mathbb{I}[x]=1$ if $x$ is true and $0$ otherwise. The threshold $\tau_{K}$ is set in a way that only the pixels with top-K highest losses are selected. We set $K$ to 5 in experiments. By calculating two losses $L_{CAE}$ and $L_{IAE}$, the pixel embeddings learn to be closer to the corresponding category and instance query embeddings simultaneously:
\begin{align}
	L_{CAE} &= L_{emb}(P, M_{c}),\\
	L_{IAE} &= L_{emb}(P, M_{s}).
\end{align}
Finally, CIAE-loss is obtained by:
\begin{align}
L_{CIAE}&=L_{CAE}+\alpha * L_{IAE}.
\label{eq:loss_emb}
\end{align}
$\alpha$ is set to 2 to balance the training losses, and we set 5 times loss for pixels with \emph{thing} classes in $L_{IAE}$.

\subsection{Category-Aware Memory Embeddings}
\label{section:memory}
Unlike one-hot encoding for classification pre-defining sub-spaces, this work proposes to determine the sub-spaces with learnable category-aware query embeddings.

In order to map pixels into category-specific sub-spaces, during the learning process, the pixel embeddings are supposed to be distinguishable from all other classes instead of only the other classes that included in the mini-batch. To this end, the query embedding of each semantic category is slowly updated during the whole learning process as a memory embedding. Specifically, assuming there are $T$ iterations in total. $M_0$ is randomly initialized for the first iteration. At the $t$-th iteration, the mean embeddings $M_c$ of classes that exist in the current mini-batch are firstly obtained. After that, we calculate an increasing updating momentum:
\begin{equation}
	\lambda_t = 1 - 0.0001 * (1-t/T)^{3}
\end{equation}
and update the memory embeddings with:
\begin{align}
	M_{t} &= L2\_norm\big[\lambda_t * M_{t-1} + (1-\lambda_t) * M_{c}\big].
\end{align}
Both $M_c$ and \emph{stuff} query embeddings in $M_s$ are replaced with $M_{t-1}$ to calculate the CIAE-loss in Eq.~\ref{eq:loss_emb}. In this way, the memory embeddings of semantic categories divide the feature space into category-specific sub-spaces. When learning to be closer to each corresponding memory embedding, the pixel embeddings are better located in target sub-spaces.

\begin{figure*}
\centering
\includegraphics[width=.92\linewidth]{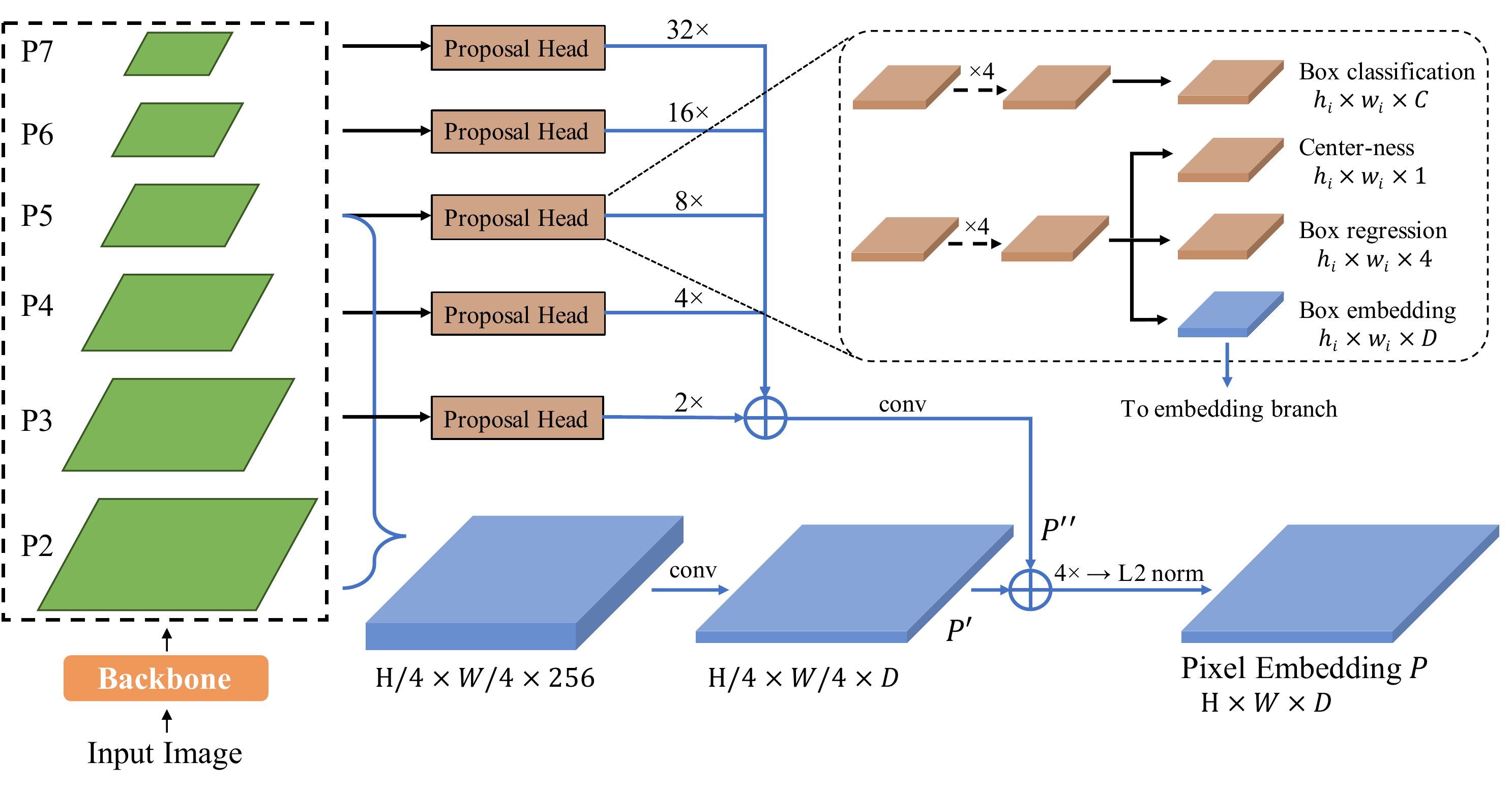}
\caption{{\bf Network structure of our proposed method.}
$H$ and $W$ are the height and width of the input image. P2 to P7 denote the feature maps of each FPN level. P3 to P7 are fed into a shared proposal head to predict bounding box classification, center-ness, box regression, and box embeddings. Features from P2 to P5 are fused as the semantic segmentation branch in Panoptic FPN~\cite{panopticFPN} for the embedding branch. All the \textcolor{blue}{\bf blue} feature maps are newly added base on FCOS~\cite{tian2019fcos:} to predict pixel embeddings. `$\oplus$' represents element-wise sum. `$4\times$' denotes 4 time bilinear up-sampling. Best viewed in color and zoom.}
\label{fig:structure}
\end{figure*}

\section{Fast Panoptic Segmentation Framework}
\subsection{Network Architecture}
As shown in Figure~\ref{fig:structure}, the proposed model consists of two branches, one for locating instances, the other for predicting pixel-wise embedding features. To locate instances, the feature maps from P3 to P7 FPN levels are fed into a shared proposal head to predict the box classification, center-ness, box regression, and box embeddings. Meanwhile, features from P2 to P5 are utilized to predict the pixel embeddings. The proposed panoptic segmentation framework is developed upon the state-of-the-art one-stage detector FCOS~\cite{tian2019fcos:}. Note that it can also be replaced with other detectors~\cite{redmon2017yolo9000,liu2016ssd:,lin2017focal}.

\noindent
{\bf Proposal head }
The proposal head is developed to locate instances. As shown in Figure~\ref{fig:structure}, two branches that consist of four convolutional layers are added after the FPN feature map for classification and regression, respectively. After that, the box classification is predicted from the classification branch while the center-ness, box regression, and embeddings are predicted from the regression branch. The box classification and box regression predict the categories and 4-directional offsets of boxes, respectively. The center-ness predicts the distances to object boundaries to down-weight the low-quality predictions during NMS. We refer readers to FCOS~\cite{tian2019fcos:} for more details. Unlike FCOS defining all inner regions of bounding boxes as training data for center-ness and box-regression, in our method, location (x,y) is considered positive only if it falls into a ground-truth mask. Different from other outputs, the box embedding is not directly supervised during training, but used to guide the pixel embedding branch, thereby enforcing the objective feature.

\noindent
{\bf Embedding branch }
\label{section:embedding_branch}
The features from P2 to P5 are utilized to predict pixel-wise embeddings, which are with the resolutions of $1/4$, $1/8$, $1/16$, and $1/32$, respectively. We employ the feature aggregation strategy similar to the semantic segmentation branch in Panoptic FPN~\cite{panopticFPN}, where features from P2 to P5 are progressively up-sampled with several $3\times3$ convolution, group normalization~\cite{wu2018group}, ReLU, and $2\times$ bilinear up-sampling (discarded for P2) operators. The up-sampled feature maps are at the same $1/4$ scale, which are then element-wise summed. In order to better encode spatial cues, CoordConv~\cite{liu2018an} is employed for the convolutions in P5. After that, an embedding map $P^{'}$ is predicted with a convolution. Instead of output the embedding map directly, we propose to enhance the multi-scale objective information by fusing box embedding predictions from each FPN level. Concretely, box embeddings from proposal heads are up-sampled and element-wise summed, following with a convolution to produce $P^{''}$. Finally, the element-wise sum of $P^{'}$ and $P^{''}$, following with $4\times$ bilinear up-sampling and L2 normalization is utilized to predict the embedding map $P$. Unless specified, our model predicts 32-dim embedding maps in practice.

\noindent
{\bf Training loss }
The total training loss is calculated as:
\begin{equation}
	L=L_{centerness}+L_{class}+L_{reg}+ L_{CIAE}.
\end{equation}
$L_{CIAE}$ is defined in Eq.~\ref{eq:loss_emb}. $L_{centerness}$, $L_{class}$, and $L_{reg}$ denote the losses for center-ness, box classification, and regression, respectively. Similar to FCOS~\cite{tian2019fcos:}, Sigmoid Focal loss~\cite{lin2017focal}, Binary Cross Entropy (BCE) loss, and IoU loss~\cite{yu2016unitbox:} are employed respectively.

\subsection{Filtering Strategy}
\label{section:target}
In this work, instance differentiating is achieved through learning category- and instance-aware pixel embeddings. When optimized with the training loss described as Eq.~\ref{eq:loss_emb}, the network learns to make each pixel embedding distinguishable from that of other instances. Previous works~\cite{discriminative_loss_2017_cvprw,deep_metric} have shown the network can be roughly trained for distinguishing pixel embeddings from any two instances. However, learning to distinguish distant instances requires a large field of view, making it difficult to optimize the network. To this end, this work proposes a remote instances filtering strategy, based on the instance bounding boxes. Concretely, the similarity between $p_i$ and $m^{k}_{s}$ is set to 0 if the pixel lies outside the bounding box of instance $k$. Ground truth and predicted bounding boxes are utilized for training and inference, respectively. Moreover, in the training phase, the ground truth boxes are expanded to 1.5 times to increase the training data and enhance the robustness of imperfect predictions.

\begin{figure*}
\centering
\includegraphics[width=.95\linewidth]{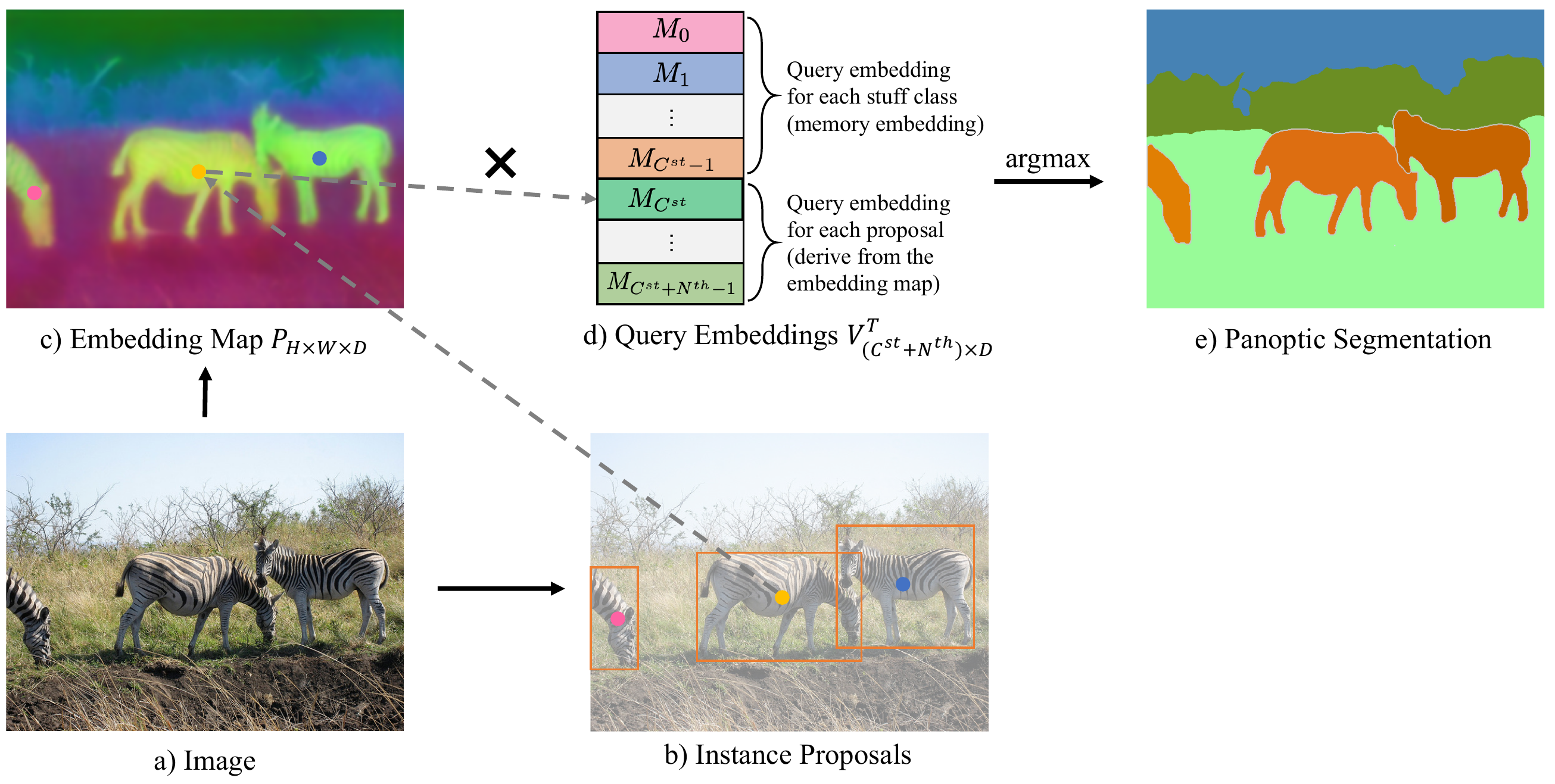}
\caption{{\bf Panoptic segmentation from pixel embeddings. }
Panoptic segmentation result is obtained from the pixel embedding map and detected instance proposals (after NMS) by calculating the cosine similarity between pixel embeddings and the query embeddings. Query embeddings consist of query embedding for each \emph{stuff} class and instance proposal. The former is the memory embedding for each \emph{stuff} class and the later is derived from the embedding map, according to where each proposal is generated. c) is the visualization result of the learned embedding map with PCA dimensionality reduction. Best viewed in color and zoom.}
\label{fig:inference}
\end{figure*}

\subsection{Inference Process}
At the inference time, the PS result is derived by assigning each pixel to a detected instance or a \emph{stuff} class according to the learned embedding. As shown in Figure~\ref{fig:inference}, given the learned embedding map $P$ and the detected instance proposals after NMS, the image is segmented by comparing the similarity between pixel embeddings $P$ and query embeddings $V$.

As illustrated in Figure~\ref{fig:inference}(d), query embeddings consist of query embedding for each \emph{stuff} class and each instance proposal. The learned memory embedding is employed as the query embedding for each \emph{stuff} class, which is consistent with the training process. For the query embedding of each proposal, we simply derive it from the embedding map, according to where each proposal is generated. Specifically, if the proposed instance is detected from the location $(x,y)$, then we use $P_{x,y}$ as the query embedding. Cosine distance is employed to measure the similarity. Since the norm of both pixel embeddings and query embeddings are 1, the cosine similarity can be simply calculated by matrix multiplication:
\begin{equation}
	P \times V^{T}=S.
\end{equation}
Each pixel is then assigned to the instance proposal (or \emph{stuff} class) with the highest similarity. After that, the semantic category of instance segments can be provided by the learned pixel embedding or box classification branch. The former can be realized by calculating the similarities between each proposal query embedding and category-aware memory embeddings of \emph{thing} classes. These two approaches are compared through experiments in Table~\ref{table:ablation_merge} (detailed in Subsection~\ref{section:ablation}). For simplicity and accuracy, we finally determine the semantic category of each instance segment with the box classification result and merge the \emph{thing} classes as a single class when learning CIAE. Moreover, \emph{stuff} regions with a size smaller than 4096 are avoided.

\section{Experiments}
\subsection{Dataset}
Our experiments are conducted on the large-scale object detection and segmentation dataset COCO~\cite{lin2014microsoft}, which contains 118K and 5K images for training and validation, respectively. The labels consist of 53 \emph{stuff} classes and 80 \emph{thing} classes that contain instance level annotation. We finally submit the result of test-dev split (20K images) to the COCO 2019 panoptic segmentation evaluation server. 

\subsection{Metric}
The results are evaluated with the standard Panoptic Quality (PQ) metric introduced by Kirillov et al.~\cite{panoptic}. 
PQ can be further divided into the multiplication of Segmentation Quality (SQ) term and Recognition Quality (RQ) term to evaluate  segmentation and recognition performances, respectively. The formulation of PQ is:
\begin{equation}
\footnotesize
	\mathrm{PQ}=\underbrace{\frac{\sum_{p,g\in TP}{\mathrm{IoU}(p,g)}}{|TP|}}_{\text{Segmentation Quality (SQ)}}\times\underbrace{\frac{|TP|}{|TP|+\frac{1}{2}|FP|+\frac{1}{2}|FN|}}_{\text{Recognition Quality (RQ)}},
\end{equation}
where $\mathrm{IoU}(p,g)$ represents the intersection-over-union between the predicted segment $p$ and ground truth segment $g$. $TP$, $FN$, and $FP$ represent matched pairs of segments ($\mathrm{IoU}(p,g)>0.5$), unmatched ground truth segments, and unmatched predicted segments, respectively. PQ, SQ, and RQ of both \emph{thing} and \emph{stuff} classes are reported in this work.

\subsection{Implementation details}
Our experiments are implemented with PyTorch~\cite{steiner2019pytorch:} and trained with 8 NVIDIA 2080Ti GPUs. The inference time is measured on 1 NVIDIA 2080Ti GPU with batch size as 1.

Unless specified, ResNet-50~\cite{he2016deep} with FPN~\cite{lin2017feature} is utilized as our backbone network. The same hyper-parameters with RetinaNet~\cite{lin2017focal} are utilized. Specifically, our network is trained for 90K iterations with stochastic gradient descent (SGD) and the mini-batch is set to 16. The learning rate is initialized to 0.01 and multiplies a factor of 0.1 at the 60K and 80K iterations, respectively. Weight decay and momentum are set as $1e^{-4}$ and 0.9, respectively. The backbone network is initialized with the ImageNet~\cite{deng2009imagenet} pre-trained weights and the newly added layers are initialized as in RetinaNet~\cite{lin2017focal}. Unless specified, we resize the shorter side of the input images to $800$ for both training and testing.

\begin{table*}
\begin{center}
\setlength{\tabcolsep}{3pt}
\caption{Results on COCO validation set. ${\dagger}$: methods with multi-scale testing. ${\ddagger}$: methods with horizontal flipping. ${\star}$: results measured with the official source code in the same environment as ours. `-512' refers to resize the shorter image side to 512 pixels. The best two results are marked in \textcolor{red}{red} and \textcolor{blue}{blue}. Results are reported as percentages.}
\label{table:valset}
  \begin{tabular}{llcccccc}
  \bottomrule
  \noalign{\smallskip}
  Method&Backbone&PQ$\uparrow$&PQ$^{th}$$\uparrow$&PQ$^{st}$$\uparrow$&Device & Inf. (ms)$\downarrow$&Fusing (ms)$\downarrow$\\
  \midrule[.8pt]
  {\it two-stage:}&&&&&&\\
  AdaptIS$^{\ddagger}$~\cite{Sofiiuk_2019_ICCV}&ResNet-50&35.9&40.3&29.3&-&-\\
  OANet~\cite{Liu_2019_CVPR}&ResNet-50-FPN&39.0&48.3&24.9&-&-&-\\
  Panoptic FPN~\cite{panopticFPN}&ResNet-50-FPN&39.4&45.9&\textcolor{blue}{\bf 29.5}&2080ti&121$^\star$&22$^\star$\\
  AUNet~\cite{Li_2019_CVPR}&ResNet-50-FPN&\textcolor{blue}{\bf 39.6}&\textcolor{red}{\bf 49.1}&25.2&-&-&-\\
  UPSNet~\cite{xiong19upsnet}&ResNet-50-FPN&\textcolor{red}{\bf 42.5}&\textcolor{blue}{\bf 48.5}&\textcolor{red}{\bf 33.4}&2080Ti&174$^\star$&22$^\star$\\
  \noalign{\smallskip}
  \hline
  \noalign{\smallskip}
  {\it one-stage:}&&&&&&\\
  DeeperLab~\cite{arxiv_2019_yang_deeperlab}&Xception-71~\cite{chollet2017xception:}&33.8&-&-&V100&119&25\\
  SSAP$^{\dagger\ddagger}$~\cite{SSAP_Gao_ICCV}&ResNet-101&36.5&-&-&-&-&-\\
  RealTimePan~\cite{real-time-panoptic}&ResNet-50-FPN&37.1&41.0&31.3&V100&63&-\\
  PCV~\cite{PCV_Wang_2020_CVPR}&ResNet-50-FPN&37.5&40.0&33.7&1080Ti&176&-\\
  Panoptic-DeepLab~\cite{cheng2020panoptic}&Xception-71&39.7&43.9&33.2&V100&132&-\\
  \bf Ours-512&ResNet-50-FPN&38.7&43.2&32.8&2080Ti&48&4\\
  \bf Ours-640&ResNet-50-FPN&39.5&44.4&33.1&2080Ti&59&6\\
  \bf Ours-800&ResNet-50-FPN&40.2&45.3&32.3&2080Ti&80&7\\
  \bf Ours-800&ResNet-101-FPN&41.4&46.4&33.9&2080Ti&96&7\\
  \bf Ours-800&ResNet-101-FPN-DCN~\cite{dai2017deformable}&\textcolor{blue}{\bf 44.2}&\textcolor{blue}{\bf 49.2}&\textcolor{blue}{\bf 36.7}&2080Ti&111&7\\
  \bf Ours-800&ResNeXt-101-FPN-DCN&\textcolor{red}{\bf 45.7}&\textcolor{red}{\bf 51.2}&\textcolor{red}{\bf 37.5} &2080Ti&176&7\\
  \bottomrule
  \end{tabular}
\end{center}
\end{table*}
\begin{table*}
\begin{center}
\setlength{\tabcolsep}{3pt}
\caption{Results on COCO test-dev set. ${\dagger}$: methods with multi-scale testing. ${\ddagger}$: methods with horizontal flipping. The best two results are marked in \textcolor{red}{red} and \textcolor{blue}{blue}. Results are reported as percentages.}
\label{table:testdevset}
  \begin{tabular}{llccccccccc}
  \bottomrule
  \noalign{\smallskip}
  Method&Backbone&PQ&SQ&RQ&PQ$^{th}$&SQ$^{th}$&RQ$^{th}$&PQ$^{st}$&SQ$^{st}$&RQ$^{st}$\\
  \midrule[.8pt]
  {\it two-stage:}&&&&&&&&\\
  Panoptic FPN~\cite{panopticFPN}&ResNet-101-FPN&40.9&-&-&48.3&-&-& 29.7&-&- \\
  OANet~\cite{Liu_2019_CVPR}&ResNet-101-FPN&41.3&-&-&50.4&-&-&27.7&-&-\\
  AdaptIS$^{\ddagger}$~\cite{Sofiiuk_2019_ICCV}&ResNeXt-101-FPN&42.8&-&-&50.1&-&-&31.8&-&-\\
  AUNet~\cite{Li_2019_CVPR}&ResNeXt-152-FPN&\textcolor{blue}{\bf 46.5}&\textcolor{red}{\bf 81.0}&\textcolor{blue}{\bf 56.1}&\textcolor{red}{\bf 55.9}&\textcolor{red}{\bf 83.7}&\textcolor{red}{\bf 66.3}&\textcolor{blue}{\bf 32.5}&\textcolor{blue}{\bf 77.0}&\textcolor{blue}{\bf 40.7} \\
  UPSNet$^{\dagger}$~\cite{xiong19upsnet}&ResNet-101-FPN-DCN&\textcolor{red}{\bf 46.6}&\textcolor{blue}{\bf 80.5}&\textcolor{red}{\bf 56.9}&\textcolor{blue}{\bf 53.2}&\textcolor{blue}{\bf 81.5}&\textcolor{blue}{\bf 64.6}&\textcolor{red}{\bf 36.7}&\textcolor{red}{\bf 78.9}&\textcolor{red}{\bf 45.3}\\
  \noalign{\smallskip}
  \hline
  \noalign{\smallskip}
  {\it one-stage:}&&&&&&&&&\\
  DeeperLab~\cite{arxiv_2019_yang_deeperlab}&Xception-71&34.3&77.1&43.1&37.5&77.5&46.8&29.6&76.4&37.4 \\
  SSAP$^{\dagger\ddagger}$~\cite{SSAP_Gao_ICCV}&ResNet-101&36.9&\textcolor{red}{\bf 80.7}&44.8&40.1&\textcolor{red}{\bf 81.6}&48.5&32.0&\textcolor{red}{\bf 79.4}&39.3\\
  PCV~\cite{PCV_Wang_2020_CVPR}&ResNet-50-FPN&37.7&77.8&47.3&40.7&78.7&50.7&33.1&76.3&42.0\\
  Panoptic-DeepLab~\cite{cheng2020panoptic}&Xception-71&39.6&-&-&-&-&-&-&-&-\\
  \bf Ours&ResNet-50-FPN&40.3&78.1&50.3&45.2&79.7&55.9&32.9&75.8&41.8\\
  \bf Ours&ResNet-101-FPN&42.1&78.7&52.3&47.6&80.2&58.5&34.0&76.4&42.9\\
  \bf Ours&ResNet-101-FPN-DCN&\textcolor{blue}{\bf 44.5}&80.0&\textcolor{blue}{\bf 54.6}&\textcolor{blue}{\bf 49.7}&81.1&\textcolor{blue}{\bf 60.6}&\textcolor{blue}{\bf 36.8}&78.3&\textcolor{blue}{\bf 45.7}\\
  \bf Ours&ResNeXt-101-FPN-DCN&\textcolor{red}{\bf 46.3}&\textcolor{blue}{\bf 80.4}&\textcolor{red}{\bf 56.6}&\textcolor{red}{\bf 51.8}&\textcolor{blue}{\bf 81.4}&\textcolor{red}{\bf 63.0}&\textcolor{red}{\bf 37.9}&\textcolor{blue}{\bf 78.8}&\textcolor{red}{\bf 46.9}\\
  \bottomrule
  \end{tabular}
\end{center}
\end{table*}

\subsection{Main Results}
\noindent
{\bf Performance on validation set }
We summarize the accuracy and speed of our method on the COCO validation set in Table~\ref{table:valset}, where the networks are trained longer (180K iterations) with scale jitter (shorter image side in [640, 800]). We compare the accuracy and efficiency of our method with the state-of-the-art two-stage and one-stage panoptic segmentation algorithms. For inference time, the average end-to-end inference time over the whole validation set is reported. The fusing time for obtaining the final PS result from multiple outputs (in our method: instance proposals and embedding map) is also reported, which is included in the inference time. Our method outperforms all the one-stage methods in both accuracy and inference speed. We are also the first one-stage method that achieves comparable performance against time-consuming two-stage methods.

\noindent
{\bf Performance on test-dev set }
The comparison with the state-of-the-art methods for panoptic segmentation on the COCO test-dev set are summarized in Table~\ref{table:testdevset}. The results of our method are predicted with a single model and single inference. No test-time augmentations such as horizontal flipping or multi-scale testing were adopted.

\begin{figure*}
  \setlength{\tabcolsep}{2pt}
  \vspace{-.5em}
  \begin{center}
  \small
  \begin{tabular}{c}
    \includegraphics[width=0.98\linewidth]{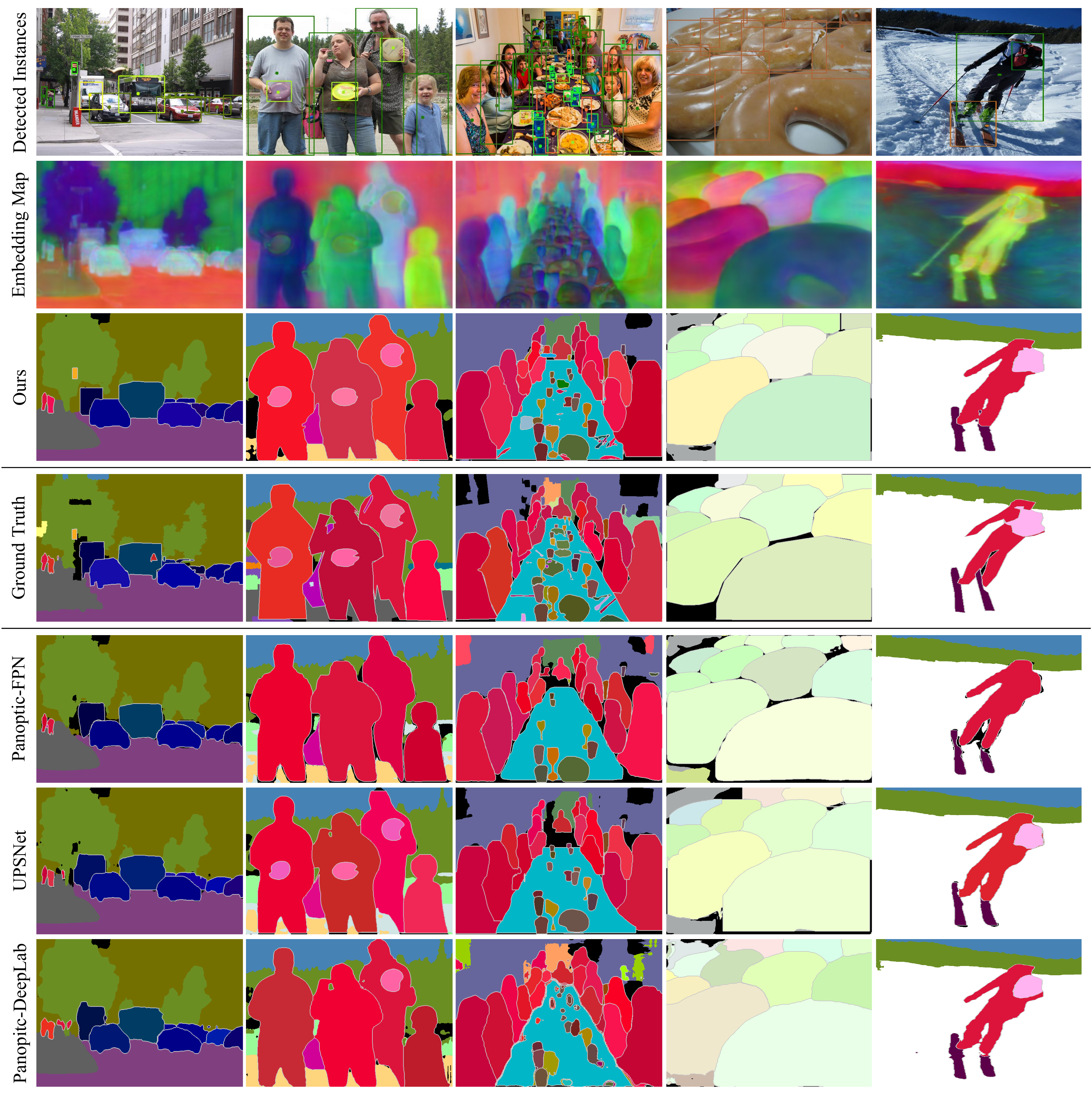}\\
  \end{tabular}
  \end{center}
    \caption{Visualization of sampled results from COCO validation set. Only single scale inference is used and our method achieves 41.4\% PQ while runs at 10 fps. 
    The first row shows detected instances. 
    The second row visualizes the learned embedding maps. 
    The third row represents the final panoptic segmentation results. 
    The fourth row shows the ground truth. 
    The last three rows show the panoptic segmentation results of other methods~\cite{panopticFPN,xiong19upsnet,cheng2020panoptic}. 
    Best viewed in color and zoom.}
  \label{fig:visualization}
\end{figure*}

\subsection{Visualization Results}
The sampled results from the COCO validation set are visualized in Figure~\ref{fig:visualization}. The first row shows the predicted instance proposals and where they are generated. The second row visualizes the learned embedding maps after PCA dimensionality reduction. The third and fourth rows represent the final panoptic segmentation results and the ground truths, respectively.
The last three rows show the results of other state-of-art methods~\cite{panopticFPN,xiong19upsnet,cheng2020panoptic}.
Only single scale inference is used and our model achieves 41.4\% PQ while runs at 10 fps. 

\subsection{Ablation Experiments}
\label{section:ablation}
\noindent
{\bf Consistently modeling \emph{stuff} classes}
The proposed PS framework models \emph{thing} and \emph{stuff} classes consistently with CIAE. To demonstrate the benefits, we conduct an experiment to segment \emph{stuff} classes with an additional semantic segmentation branch, where an $H/4 \times W/4 \times (Nst + 1)$ tensor (\emph{thing} classes are merged as a single class) is predicted in parallel with $P'$ and supervised with the standard cross-entropy loss. The results are shown in Table~\ref{table:ablation_fcnstuff}. Similar performances are achieved in mIoU and PQ$^{st}$, however, the performance for PQ$^{th}$ is much lower. This is because the consistent modeling leads to more precise boundaries, which is especially beneficial for small instances.

\begin{table}
  \begin{center}
  \setlength{\tabcolsep}{4pt}
    \caption{Influence of consistently modeling stuff classes with CIAE. The first row shows the performance of separately modeling stuff classes with a FCN-style semantic segmentation branch.}
    \vspace{-.5em}
    \label{table:ablation_fcnstuff}
    \begin{tabular}{ccccc}
    \bottomrule
    \noalign{\smallskip}
    Modeling of \emph{stuff} classes&PQ (\%)&PQ$^{th}$ (\%)&PQ$^{st}$ (\%)&mIoU (\%)\\
    \midrule[.8pt]
    Inconsistent &35.6&38.1&30.6&48.0\\
    Consistent   &37.9&42.5&30.9&48.5\\
    Improve.     &+2.3&+4.4&+0.3&+0.5\\ 
    \bottomrule
    \end{tabular}
  \end{center}
  \vspace{.5em}
  \begin{center}
  \setlength{\tabcolsep}{2.5pt}
    \caption{Influence of deriving proposal categories from CIAE or detector. 
    `E' and `D' indicate to derive instance categories from the embedding map and box classification branch, respectively.
    `CateNumE' refers to the predicted category number of the embedding map.
    `CateNumD' denotes the predicted category number of the detector.}
    \vspace{-.5em}
    \label{table:ablation_merge}
    \begin{tabular}{lllcccc}
    \bottomrule
    \noalign{\smallskip}
    &CateNumE &CateNumD &PQ (\%)&PQ$^{th}$ (\%)&PQ$^{st}$ (\%)&mIoU (\%)\\
    \midrule[.8pt]
    E$\quad$ &$C^{st}+C^{th}$ &$1+C^{th}$ &36.7&40.7&30.4&48.8\\
    D        &$C^{st}+C^{th}$ &$1+C^{th}$ &37.7&\bf{42.6}&30.4&\bf{49.9}\\
    \noalign{\smallskip}
    \hline
    \noalign{\smallskip}
    E        &$C^{st}+C^{th}$ &$2$        &35.5&39.4&29.6&47.5\\ 
    D        &$C^{st}+1$      &$1+C^{th}$ &\bf{37.9}&42.5&\bf{30.9}&48.5\\
    \noalign{\smallskip}
    \hline
    \noalign{\smallskip}
    D        &$C^{st}$        &$1+C^{th}$ &37.1&41.8&30.1&48.4\\
    \bottomrule
    \end{tabular}
  \end{center}
  \begin{center}
    \setlength{\tabcolsep}{6.5pt}
    \caption{Oracle performance. Ground truth bounding boxes and masks are employed. Query embedding for each instance is calculated as the mean embedding.}
    \vspace{-.3em}
    \label{table:ablation_oracle}
    \begin{tabular}{lcccc}
    \bottomrule
    \noalign{\smallskip}
    Backbone&Oracle&PQ (\%)&PQ$^{th}$ (\%)&PQ$^{st}$ (\%)\\
    \midrule[.8pt]
    ResNet-50-FPN&&37.9&42.5&30.9\\    
    ResNet-50-FPN&\checkmark&59.2&75.8&34.1\\
    \noalign{\smallskip}
    \hline
    \noalign{\smallskip}
    ResNet-101-FPN&&40.8&45.4&33.8\\
    ResNet-101-FPN&\checkmark&60.0&76.2&35.7\\
    \bottomrule
    \end{tabular}
  \end{center}
\end{table}

\noindent
{\bf Proposal categories from CIAE or detector}
In this part, we test the difference between deriving instance categories through the learned embedding map or the box classification branch. We start by learning specific \emph{thing} classes with both branches and inferencing with each of them. As shown in the first two rows in Table~\ref{table:ablation_merge}, the box classification branch classifies instance proposals better than the embedding map (37.7\% \emph{vs.} 36.7\%). After that, the two branches separately merge \emph{thing} classes as a single class and infer from the other. Experiments show that employing a class-agnostic box classifier reduces PQ by 1.2\%. However, since the information to be encoded in CIAE is reduced, merging \emph{thing} classes in learning CIAE promotes PQ$^{st}$ with 0.5\% and obtains slightly better performance. This setting is used as the default because of its accuracy and simplicity, where the \emph{thing} classes are merged as a single class when learning CIAE and the specific category of each instance segment is determined with the box classification result. Moreover, from the last two rows in Table~\ref{table:ablation_merge}, we find that if the category information for \emph{thing} classes is completely abandoned (realized by setting $L_{CAE}$ to 0) in CIAE, PQ degrades with 0.8\%. This result supports our motivation to encode pixel-wise semantic-classification and instance-distinction simultaneously. Specifically, by adding the CAE loss, the model learns CIAE instead of IAE, and locates the embedding vectors for \emph{thing} instances into a specific subspace, thereby making the network easier for optimization, and finally boosts the performance for both \emph{thing} and \emph{stuff} classes consistently. Importantly, the improvements are achieved only by adding a constraint during training, which does not introduce additional parameters and can boost the performance at no cost. 

\noindent
{\bf Oracle performance}
We test the upper limits of the proposed method by employing predicted pixel embeddings and ground-truth (GT) bounding boxes for inference.
The query embedding of each instance is derived by calculating the mean embedding within the GT mask. The results are shown in Table~\ref{table:ablation_oracle}. Employing GT bounding boxes nearly doubles the performance of \emph{thing} classes, which shows the great potential of the proposed pixel embedding based framework.

\begin{figure}
  \setlength{\tabcolsep}{1pt}
  \begin{center}
  \begin{tabular}{c}
    \includegraphics[width=0.98\linewidth]{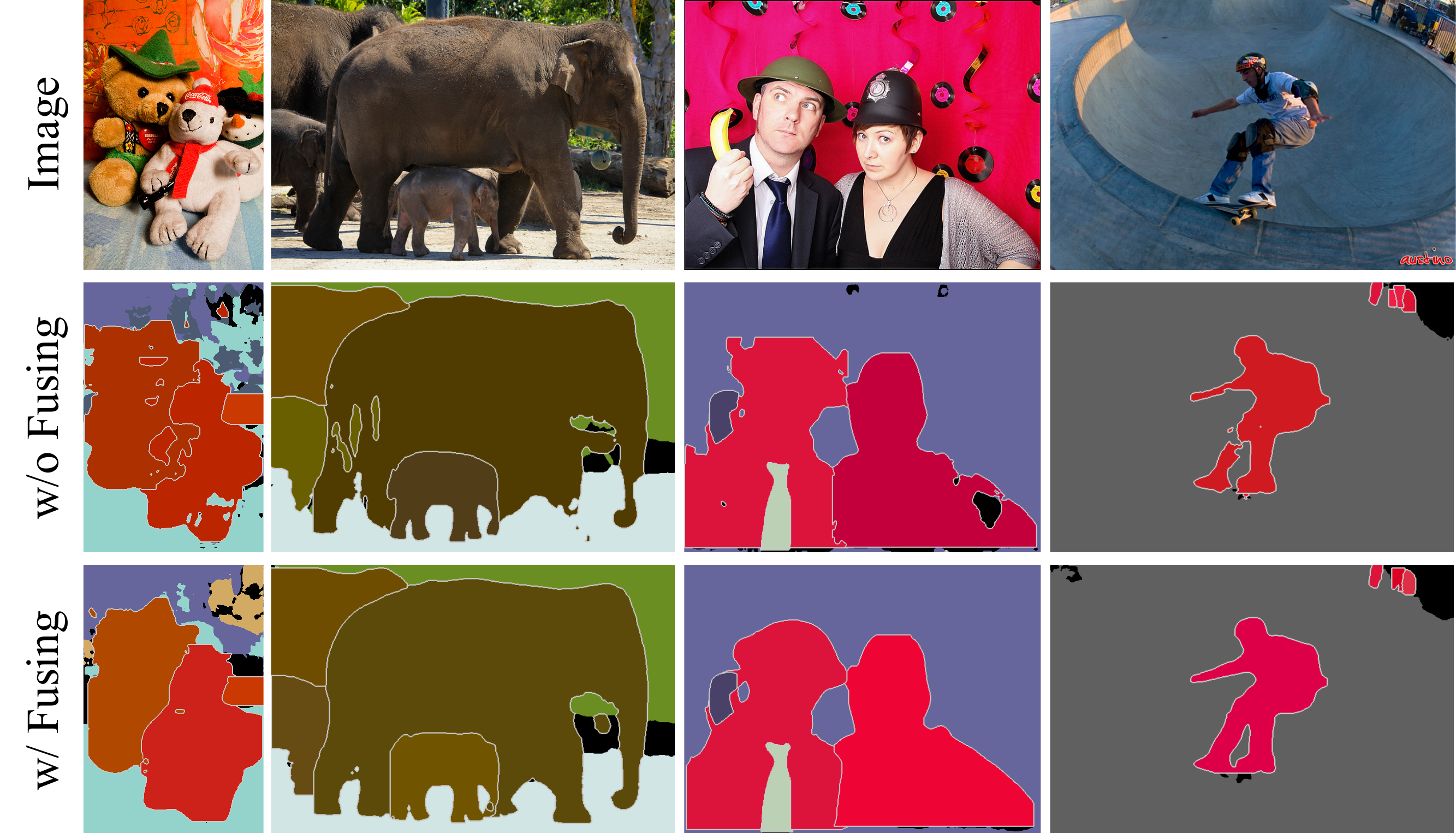}\\
  \end{tabular}
  \end{center}
\caption{Visualization of predictions with or without fusing box embeddings. Fusing box embeddings to the embedding branch helps to produce more smooth and complete instance segments. Best viewed in color and zoom.}
\label{fig:vis_fuse}
\end{figure}
\begin{table}
\begin{center}
  \setlength{\tabcolsep}{5pt}
  \vspace{-.3em}
    \caption{Influence of employing CoordConv and fusing box embeddings to the embedding branch.}
    \label{table:ablation_fea}
    \begin{tabular}{ccccc}
    \bottomrule
    \noalign{\smallskip}
    CoordConv&Box Embedding&PQ (\%)&PQ$^{th}$ (\%)&PQ$^{st}$ (\%)\\
    \midrule[.8pt]
              &          &37.1&41.8&30.0\\
    \checkmark&          &37.5&42.3&30.2\\
              &\checkmark&37.8&\bf{42.5}&30.8\\
    \checkmark&\checkmark&\bf{37.9}&\bf{42.5}&\bf{30.9}\\
    \bottomrule
    \end{tabular}
  \end{center}
  \begin{center}
    \setlength{\tabcolsep}{6pt}
    \caption{Results of using different memory updating momentum.}
    \label{table:ablation_factor}
    \begin{tabular}{lccc}
      \bottomrule
      \noalign{\smallskip}
      Momentum&PQ (\%)&PQ$^{th}$ (\%)&PQ$^{st}$ (\%)\\
      \midrule[.8pt]
      0&\multicolumn{3}{c}{fail to converge}\\
      0.99&37.5&42.2&30.2\\
      0.999&37.7&\bf{42.6}&30.3\\
      0.9999&\bf{37.9}&42.5&\bf{30.9}\\
      0.99999&37.4&42.1&30.4\\
      \bottomrule
    \end{tabular}
  \end{center}
  \vspace{-0.5em}
\end{table}

\noindent
{\bf Box embedding and CoordConv}
As described in Subsection~\ref{section:embedding_branch}, we propose employing CoordConv and fusing the box embeddings from each FPN level to the embedding branch to enhance the objective information. Both quantitative and qualitative analysis of this structure is carried out. The proposed box embedding structure is compared with the one where the box embedding branch is abandoned. In other words, $P=f(P^{'} + P^{''})$ is replaced with $P=f(P^{'})$, where $f$ denotes $4\times$ up-sampling and L2 normalization. As shown in Table~\ref{table:ablation_fea}, fusing box embeddings boosts the performance from 37.1\% to 37.8\% PQ. Both performances of \emph{thing} and \emph{stuff} classes are improved. The visualization results in Figure~\ref{fig:vis_fuse} indicate that more smooth and complete instance segments are generated after fusing the box embeddings to the embedding branch.

\noindent
{\bf Memory updating momentum}
As described in Subsection~\ref{section:memory}, the category-aware memory embeddings are gradually updated and finally employed as query embeddings at the inference process. We set a large momentum value to stabilize the training process. As shown in Table~\ref{table:ablation_factor}, the network learns well with a large momentum value. When the momentum is set to 0, the training loss oscillates and fails to converge.

\noindent
{\bf Filtering strategy }
As described in Subsection~\ref{section:target}, this work proposes a box based filtering strategy to eliminate distant instances from optimizing, and the ground truth boxes are expanded during training. As shown in Table~\ref{table:ablation_samp}, expanding the ground truth boxes to 1.5 times achieves the best performance. The first row shows the performance drops with 1.3\% PQ when the filtering strategy is abandoned, which represents the similarities to all instance mean embeddings are optimized during training.

\noindent
{\bf Triplet loss margin}
This work proposes to learn CIAE by optimizing the triplet losses with category and instance query embeddings. Here we test the performances for different margins. Note that the margins in $L_{CAE}$ and $L_{IAE}$ can be different, but we set them the same for simplifying. As shown in Table~\ref{table:ablation_margin}, the model performs best with $margin=0.15$. It is difficult to optimize the network with an excessive large margin, and the learned embeddings lack discrimination when the margin is too small. 

\noindent
{\bf Embedding dimension }
Table~\ref{table:ablation_dim} shows the results of using different embedding dimensions. We found the embedding dimension only slightly affects the performance for \emph{stuff} classes, and the model works well even trained with 16 dimension embeddings.

\begin{table}
\begin{center}
  \setlength{\tabcolsep}{8.5pt}
    \caption{Influence of the filtering strategy. The first row shows the performance of abandoning filtering strategy.}
    \label{table:ablation_samp}
    \begin{tabular}{cccc}
      \bottomrule
      \noalign{\smallskip}
      Box&PQ (\%)&PQ$^{th}$ (\%)&PQ$^{st}$ (\%)\\
      \midrule[.8pt]
           All    &36.6&41.1&29.8\\
      2.0$\times$ &37.4&42.1&30.4\\
      1.5$\times$ &\bf{37.9}&\bf{42.5}&\bf{30.9}\\
      1.2$\times$ &37.8&42.3&30.8\\
      1.0$\times$ &36.6&40.8&30.1\\
      \bottomrule
    \end{tabular}
  \end{center}
  \begin{center}
    \setlength{\tabcolsep}{8.pt}
    \caption{Results of using different triplet loss margins.}
    \label{table:ablation_margin}
    \begin{tabular}{cccc}
      \bottomrule
      \noalign{\smallskip}
      Margin&PQ (\%)&PQ$^{th}$ (\%)&PQ$^{st}$ (\%)\\
      \midrule[.8pt]
      0.30&36.5&39.9&30.9\\
      0.20&37.7&41.9&\bf{31.0}\\
      0.15&\bf{37.9}&\bf{42.5}&30.9\\
      0.10&37.6&42.5&30.2\\
      0.00&28.7&36.2&17.2\\
      \bottomrule
    \end{tabular}
  \end{center}
\vspace{.0em}
  \begin{center}
    \caption{Results of using different embedding dimensions.}
    \setlength{\tabcolsep}{8.5pt}
    \label{table:ablation_dim}
    \begin{tabular}{cccc}
      \bottomrule
      \noalign{\smallskip}
      Dim&PQ (\%)&PQ$^{th}$ (\%)&PQ$^{st}$ (\%)\\
      \midrule[.8pt]
      48&\bf{37.9}&42.5&\bf{31.0}\\
      32&\bf{37.9}&42.5&30.9\\
      24&37.8&\bf{42.6}&30.6\\
      16&37.3&42.2&30.0\\
      \bottomrule
    \end{tabular}
  \end{center}
\end{table}

\section{Conclusion}
\label{section:conlusion}
In this paper, we have introduced a novel one-stage panoptic segmentation method that consistently models \emph{thing} and \emph{stuff} classes. Our method is based on incorporating a one-stage detector with the proposed category- and instance-aware pixel embedding (CIAE). CIAE is a pixel-wise embedding feature that encodes both semantic-classification and instance-distinction information. The panoptic segmentation result is simply predicted by assigning each pixel to a detected instance or a \emph{stuff} class according to the learned embedding. Our method has a fast runtime speed and is also the first one-stage method that achieves comparable performance against two-stage methods on the challenging COCO benchmark.


\begin{IEEEbiography}[{\includegraphics[height=1.2in,clip,keepaspectratio]{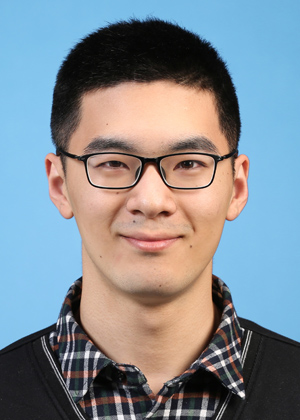}}]{Naiyu Gao}
received the B.Eng. degree in computer electronic and information engineering from Xi’an Jiaotong University (XJTU), Xi’an, China, in 2017. He is currently pursuing the Ph.D. degree with the Institute of Automation, Chinese Academy of Sciences (CASIA), Beijing, China. 
His research interests include image segmentation, deep learning, pattern recognition, and computer vision.
\end{IEEEbiography}
\begin{IEEEbiography}
[{\includegraphics[height=1.2in,clip,keepaspectratio]{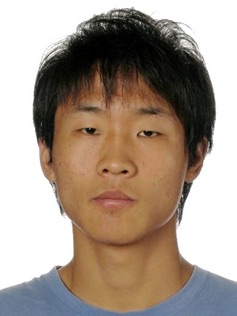}}]{Yanhu Shan} received the B.S. degree from
Beijing Information Science and Technology University, Beijing, China, in 2009 and the Ph.D. degree in pattern recognition and intelligent systems from the National Laboratory of Pattern Recognition, Institute of Automation, Chinese Academy of Sciences, Beijing, China, in 2015. He is a researcher at Horizon Robotics from 2017. He was formerly with Samsung R\&D Institute after his graduation. His research interests include compute vision and deep learning.
\end{IEEEbiography}
\begin{IEEEbiography}[{\includegraphics[height=1.2in,clip,keepaspectratio]{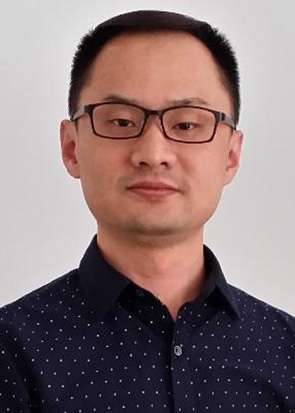}}]{Xin Zhao}
received the Ph.D. degree from the University of Science and Technology of China. He is currently an Associate Professor in the Institute of Automation, Chinese Academy of Sciences (CASIA). His current research interests include pattern recognition, computer vision, and machine learning. He received the International Association of Pattern Recognition Best Student Paper Award at ACPR 2011. He received the 2nd place entry of COCO Panoptic Challenge at ECCV 2018. 
\end{IEEEbiography}
\begin{IEEEbiography}[{\includegraphics[height=1.2in,clip,keepaspectratio]{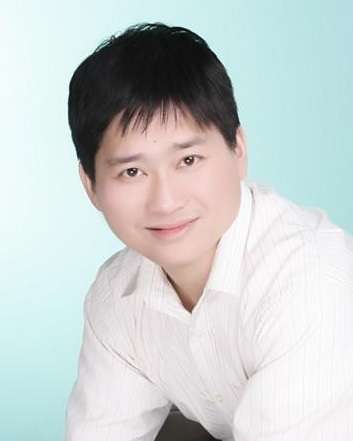}}]{Kaiqi Huang} received the B.Sc. and M.Sc. degrees from the Nanjing University of Science Technology, China, and the Ph.D. degree from Southeast University. He is a Full Professor in the Center for Research on Intelligent System and Engineering (CRISE), Institute of Automation, Chinese Academy of Sciences (CASIA). He is also with the University of Chinese Academy of Sciences (UCAS), and the CAS Center for Excellence in Brain Science and Intelligence Technology. He has published over 210 papers in the important international journals and conferences, such as the IEEE TPAMI, T-IP, T-SMCB, TCSVT, Pattern Recognition, CVIU, ICCV, ECCV, CVPR, ICIP, and ICPR. His current researches focus on computer vision,pattern recognition and game theory, including object recognition, video analysis, and visual surveillance. He serves as co-chairs and program committee members over 40 international conferences, such as ICCV, CVPR, ECCV, and the IEEE workshops on visual surveillance. He is an Associate Editor of the IEEE TRANSACTIONS ON SYSTEMS, MAN, AND CYBERNETICS: SYSTEMS and Pattern Recognition.
\end{IEEEbiography}


\begin{thebibliography}{10}
\providecommand{\url}[1]{#1}
\csname url@samestyle\endcsname
\providecommand{\newblock}{\relax}
\providecommand{\bibinfo}[2]{#2}
\providecommand{\BIBentrySTDinterwordspacing}{\spaceskip=0pt\relax}
\providecommand{\BIBentryALTinterwordstretchfactor}{4}
\providecommand{\BIBentryALTinterwordspacing}{\spaceskip=\fontdimen2\font plus
\BIBentryALTinterwordstretchfactor\fontdimen3\font minus
  \fontdimen4\font\relax}
\providecommand{\BIBforeignlanguage}[2]{{%
\expandafter\ifx\csname l@#1\endcsname\relax
\typeout{** WARNING: IEEEtran.bst: No hyphenation pattern has been}%
\typeout{** loaded for the language `#1'. Using the pattern for}%
\typeout{** the default language instead.}%
\else
\language=\csname l@#1\endcsname
\fi
#2}}
\providecommand{\BIBdecl}{\relax}
\BIBdecl

\bibitem{panoptic}
A.~Kirillov, K.~He, R.~Girshick, C.~Rother, and P.~Dollar, ``Panoptic
  segmentation,'' in \emph{CVPR}, 2019.

\bibitem{panopticFPN}
A.~Kirillov, R.~Girshick, K.~He, and P.~Dollar, ``Panoptic feature pyramid
  networks,'' in \emph{CVPR}, 2019.

\bibitem{Liu_2019_CVPR}
H.~Liu, C.~Peng, C.~Yu, J.~Wang, X.~Liu, G.~Yu, and W.~Jiang, ``An end-to-end
  network for panoptic segmentation,'' in \emph{CVPR}, 2019.

\bibitem{Li_2019_CVPR}
Y.~Li, X.~Chen, Z.~Zhu, L.~Xie, G.~Huang, D.~Du, and X.~Wang,
  ``Attention-guided unified network for panoptic segmentation,'' in
  \emph{CVPR}, 2019.

\bibitem{xiong19upsnet}
Y.~Xiong, R.~Liao, H.~Zhao, R.~Hu, M.~Bai, E.~Yumer, and R.~Urtasun,
  ``{UPSNet}: A unified panoptic segmentation network,'' in \emph{CVPR}, 2019.

\bibitem{arxiv_2019_yang_deeperlab}
T.-J. Yang, M.~D. Collins, Y.~Zhu, J.-J. Hwang, T.~Liu, X.~Zhang, V.~Sze,
  G.~Papandreou, and L.-C. Chen, ``{DeeperLab}: Single-shot image parser,''
  \emph{arXiv:1902.05093}, 2019.

\bibitem{SSAP_Gao_ICCV}
N.~Gao, Y.~Shan, Y.~Wang, X.~Zhao, Y.~Yu, M.~Yang, and K.~Huang, ``{SSAP}:
  Single-shot instance segmentation with affinity pyramid,'' in \emph{ICCV},
  2019.

\bibitem{SSAP_Gao_CSVT}
N.~Gao, Y.~Shan, Y.~Wang, X.~Zhao, and K.~Huang, ``{SSAP}: Single-shot instance
  segmentation with affinity pyramid,'' \emph{IEEE TCSVT}, vol.~31, no.~2, pp.
  661--673, 2021.

\bibitem{PCV_Wang_2020_CVPR}
H.~Wang, R.~Luo, M.~Maire, and G.~Shakhnarovich, ``Pixel consensus voting for
  panoptic segmentation,'' in \emph{CVPR}, 2020.

\bibitem{cheng2020panoptic}
B.~Cheng, M.~D. Collins, Y.~Zhu, T.~Liu, T.~S. Huang, H.~Adam, and L.-C. Chen,
  ``{Panoptic-DeepLab}: A simple, strong, and fast baseline for bottom-up
  panoptic segmentation,'' in \emph{CVPR}, 2020.

\bibitem{real-time-panoptic}
R.~Hou, J.~Li, A.~Bhargava, A.~Raventos, V.~Guizilini, C.~Fang, J.~Lynch, and
  A.~Gaidon, ``Real-time panoptic segmentation from dense detections,'' in
  \emph{CVPR}, 2020.

\bibitem{discriminative_loss_2017_cvprw}
B.~D. Brabandere, D.~Neven, and L.~V. Gool, ``Semantic instance segmentation
  with a discriminative loss function,'' \emph{arXiv:1708.02551}, 2017.

\bibitem{deep_metric}
A.~Fathi, Z.~Wojna, V.~Rathod, P.~Wang, H.~O. Song, S.~Guadarrama, and K.~P.
  Murphy, ``Semantic instance segmentation via deep metric learning,''
  \emph{arXiv:1703.10277}, 2017.

\bibitem{kong2018recurrent}
S.~Kong and C.~C. Fowlkes, ``Recurrent pixel embedding for instance grouping,''
  in \emph{CVPR}, 2018.

\bibitem{lin2014microsoft}
T.~Lin, M.~Maire, S.~J. Belongie, J.~Hays, P.~Perona, D.~Ramanan, P.~Dollar,
  and C.~L. Zitnick, ``{Microsoft COCO}: Common objects in context,'' in
  \emph{ECCV}, 2014.

\bibitem{He_2017_ICCV}
K.~He, G.~Gkioxari, P.~Dollar, and R.~Girshick, ``{Mask R-CNN},'' in \emph{ICCV},
  2017.

\bibitem{ren2015faster}
S.~Ren, K.~He, R.~Girshick, and J.~Sun, ``{Faster R-CNN}: Towards real-time
  object detection with region proposal networks,'' in \emph{NIPS}, 2015.

\bibitem{Liu_2018_CVPR}
S.~Liu, L.~Qi, H.~Qin, J.~Shi, and J.~Jia, ``Path aggregation network for
  instance segmentation,'' in \emph{CVPR}, 2018.

\bibitem{chen2018masklab:}
L.~Chen, A.~Hermans, G.~Papandreou, F.~Schroff, P.~Wang, and H.~Adam,
  ``{MaskLab}: Instance segmentation by refining object detection with semantic
  and direction features,'' in \emph{CVPR}, 2018.

\bibitem{huang2019mask}
Z.~Huang, L.~Huang, Y.~Gong, C.~Huang, and X.~Wang, ``Mask scoring r-cnn,'' in
  \emph{CVPR}, 2019.

\bibitem{chen2019hybrid}
K.~Chen, W.~Ouyang, C.~C. Loy, D.~Lin, J.~Pang, J.~Wang, Y.~Xiong, X.~Li,
  S.~Sun, W.~Feng \emph{et~al.}, ``Hybrid task cascade for instance
  segmentation,'' in \emph{CVPR}, 2019.

\bibitem{xu2019explicit}
W.~Xu, H.~Wang, F.~Qi, and C.~Lu, ``Explicit shape encoding for real-time
  instance segmentation,'' in \emph{ICCV}, 2019.

\bibitem{xie2019polarmask}
E.~Xie, P.~Sun, X.~Song, W.~Wang, X.~Liu, D.~Liang, C.~Shen, and P.~Luo,
  ``{PolarMask}: Single shot instance segmentation with polar representation,''
  in \emph{CVPR}, June 2020.

\bibitem{Bolya_2019_ICCV}
D.~Bolya, C.~Zhou, F.~Xiao, and Y.~J. Lee, ``{YOLACT}: Real-time instance
  segmentation,'' in \emph{ICCV}, 2019.

\bibitem{Chen_2019_ICCV}
X.~Chen, R.~Girshick, K.~He, and P.~Dollar, ``{TensorMask}: A foundation for
  dense object segmentation,'' in \emph{ICCV}, 2019.

\bibitem{redmon2017yolo9000}
J.~Redmon and A.~Farhadi, ``{YOLO9000}: Better, faster, stronger,'' in
  \emph{CVPR}, 2017.

\bibitem{tian2019fcos:}
Z.~Tian, C.~Shen, H.~Chen, and T.~He, ``{FCOS}: Fully convolutional one-stage
  object detection,'' in \emph{ICCV}, 2019.

\bibitem{dai2016instance-aware}
J.~Dai, K.~He, and J.~Sun, ``Instance-aware semantic segmentation via
  multi-task network cascadeds,'' in \emph{CVPR}, 2016.

\bibitem{li2017fully}
Y.~Li, H.~Qi, J.~Dai, X.~Ji, and Y.~Wei, ``Fully convolutional instance-aware
  semantic segmentation,'' in \emph{CVPR}, 2017.

\bibitem{Liu_2018_ECCV}
Y.~Liu, S.~Yang, B.~Li, W.~Zhou, J.~Xu, H.~Li, and Y.~Lu, ``Affinity derivation
  and graph merge for instance segmentation,'' in \emph{ECCV}, 2018.

\bibitem{kendall2018multi-task}
A.~Kendall, Y.~Gal, and R.~Cipolla, ``Multi-task learning using uncertainty to
  weigh losses for scene geometry and semantics,'' in \emph{CVPR}, 2018.

\bibitem{Neven_2019_CVPR}
D.~Neven, B.~D. Brabandere, M.~Proesmans, and L.~V. Gool, ``Instance
  segmentation by jointly optimizing spatial embeddings and clustering
  bandwidth,'' in \emph{CVPR}, 2019.

\bibitem{fukunaga1975the}
K.~Fukunaga and L.~Hostetler, ``The estimation of the gradient of a density
  function, with applications in pattern recognition,'' \emph{IEEE Transactions
  on Information Theory}, vol.~21, no.~1, pp. 32--40, 1975.

\bibitem{Arthur2007k}
D.~Arthur and S.~Vassilvitskii, ``{k-Means++}: The advantages of carefull
  seeding,'' in \emph{Symposiumon on Discrete algorithms}, 2007.

\bibitem{porzi2019seamless}
L.~Porzi, S.~R. Bulo, A.~Colovic, and P.~Kontschieder, ``Seamless scene
  segmentation,'' in \emph{CVPR}, 2019.

\bibitem{lin2017feature}
T.~Lin, P.~Dollar, R.~Girshick, K.~He, B.~Hariharan, and S.~Belongie, ``Feature
  pyramid networks for object detection,'' in \emph{CVPR}, 2017.

\bibitem{chen2018deeplabv2:}
L.~Chen, G.~Papandreou, I.~Kokkinos, K.~P. Murphy, and A.~L. Yuille,
  ``{DeepLab}: Semantic image segmentation with deep convolutional nets, atrous
  convolution, and fully connected crfs,'' \emph{IEEE TPAMI}, vol.~40, no.~4, 2018.

\bibitem{Sofiiuk_2019_ICCV}
K.~Sofiiuk, O.~Barinova, and A.~Konushin, ``{AdaptIS}: Adaptive instance
  selection network,'' in \emph{ICCV}, 2019.

\bibitem{Lazarow_2020_CVPR}
J.~Lazarow, K.~Lee, K.~Shi, and Z.~Tu, ``Learning instance occlusion for
  panoptic segmentation,'' in \emph{CVPR}, 2020.

\bibitem{Wu_2020_CVPR}
Y.~Wu, G.~Zhang, Y.~Gao, X.~Deng, K.~Gong, X.~Liang, and L.~Lin,
  ``Bidirectional graph reasoning network for panoptic segmentation,'' in
  \emph{CVPR}, 2020.

\bibitem{Chen_2020_CVPR}
Y.~Chen, G.~Lin, S.~Li, O.~Bourahla, Y.~Wu, F.~Wang, J.~Feng, M.~Xu, and X.~Li,
  ``Banet: Bidirectional aggregation network with occlusion handling for
  panoptic segmentation,'' in \emph{CVPR}, 2020.

\bibitem{Li_2020_CVPR}
Q.~Li, X.~Qi, and P.~H. Torr, ``Unifying training and inference for panoptic
  segmentation,'' in \emph{CVPR}, 2020.

\bibitem{Papandreou_2018_ECCV}
G.~Papandreou, T.~Zhu, L.-C. Chen, S.~Gidaris, J.~Tompson, and K.~Murphy,
  ``{PersonLab}: Person pose estimation and instance segmentation with a
  bottom-up, part-based, geometric embedding model,'' in \emph{ECCV}, 2018.

\bibitem{keuper2015efficient}
M.~Keuper, E.~Levinkov, N.~Bonneel, G.~Lavoue, T.~Brox, and B.~Andres,
  ``Efficient decomposition of image and mesh graphs by lifted multicuts,'' in
  \emph{ICCV}, 2015.

\bibitem{liu2016ssd:}
W.~Liu, D.~Anguelov, D.~Erhan, C.~Szegedy, S.~E. Reed, C.~Fu, and A.~C. Berg,
  ``{SSD}: Single shot multibox detector,'' in \emph{ECCV}, 2016.

\bibitem{lin2017focal}
T.~Lin, P.~Goyal, R.~B. Girshick, K.~He, and P.~Dollar, ``Focal loss for dense
  object detection,'' in \emph{ICCV}, 2017.

\bibitem{wu2018group}
Y.~Wu and K.~He, ``Group normalization.'' in \emph{ECCV}, 2018.

\bibitem{liu2018an}
R.~Liu, J.~Lehman, P.~Molino, F.~P. Such, E.~Frank, A.~Sergeev, and
  J.~Yosinski, ``An intriguing failing of convolutional neural networks and the
  coordconv solution,'' in \emph{NIPS}, 2018.

\bibitem{yu2016unitbox:}
J.~Yu, Y.~Jiang, Z.~Wang, Z.~Cao, and T.~S. Huang, ``{UnitBox}: An advanced
  object detection network,'' in \emph{ACM Multimedia}, 2016.

\bibitem{steiner2019pytorch:}
B.~Steiner, Z.~Devito, S.~Chintala, S.~Gross, A.~Paszke, F.~Massa, A.~Lerer,
  G.~Chanan, Z.~Lin, E.~Yang \emph{et~al.}, ``Pytorch: An imperative style,
  high-performance deep learning library,'' in \emph{NIPS}, 2019.

\bibitem{he2016deep}
K.~He, X.~Zhang, S.~Ren, and J.~Sun, ``Deep residual learning for image
  recognition,'' in \emph{CVPR}, 2016.

\bibitem{deng2009imagenet}
J.~Deng, W.~Dong, R.~Socher, L.-J. Li, K.~Li, and L.~Fei-Fei, ``Imagenet: A
  large-scale hierarchical image database,'' in \emph{CVPR}, 2009.

\bibitem{chollet2017xception:}
F.~Chollet, ``Xception: Deep learning with depthwise separable convolutions,''
  in \emph{CVPR}, 2017.

\bibitem{dai2017deformable}
J.~Dai, H.~Qi, Y.~Xiong, Y.~Li, G.~Zhang, H.~Hu, and Y.~Wei, ``Deformable
  convolutional networks,'' in \emph{ICCV}, 2017.

\end{thebibliography}
\end{document}